\documentclass{article}

\usepackage{microtype}
\usepackage{graphicx}
\usepackage{subfigure}
\usepackage{booktabs} 
\usepackage{amssymb}
\usepackage{amsmath} 

\usepackage[]{soul}
\usepackage{textcomp}
\usepackage{multicol}

\usepackage{mathtools}

\usepackage{hyperref}

\usepackage[accepted]{icml2021}

\icmltitlerunning{CNN Interpretability with General Pattern Theory}

\begin{document}

\twocolumn[
\icmltitle{Convolutional Neural Network Interpretability with General Pattern Theory}



\icmlsetsymbol{equal}{*}

\begin{icmlauthorlist}
\icmlauthor{Erico Tjoa}{to}
\icmlauthor{Guan Cuntai}{to,goo}
\end{icmlauthorlist}

\icmlaffiliation{to}{Nanyang Technological University, Singapore}
\icmlaffiliation{goo}{Alibaba Inc, Hangzhou, China}

\icmlcorrespondingauthor{Guan Cuntai}{ctguan@ntu.edu.sg}
\icmlcorrespondingauthor{Erico Tjoa}{ericotjoa@gmail.com}

\icmlkeywords{Machine Learning, Deep Neural Network, Explainable Artificial Intelligence, XAI, Intepretability, Pattern Theory}

\vskip 0.3in
]


\printAffiliationsAndNotice{}  

\begin{abstract}
Ongoing efforts to understand deep neural networks (DNN) have provided many insights, but DNNs remain incompletely understood. Improving DNN's interpretability has practical benefits, such as more accountable usage, better algorithm maintenance and improvement. The complexity of dataset structure may contribute to the difficulty in solving interpretability problem arising from DNN's black-box mechanism. Thus, we propose to use pattern theory formulated by Ulf Grenander, in which data can be described as configurations of fundamental objects that allow us to investigate convolutional neural network's (CNN) interpretability in a component-wise manner. Specifically, U-Net-like structure is formed by attaching expansion blocks (EB) to ResNet, allowing it to perform semantic segmentation-like tasks at its EB output channels designed to be compatible with pattern theory's configurations. Through these modules, some heatmap-based explainable artificial intelligence (XAI) methods will be shown to extract explanations w.r.t individual generators that make up a single data sample, potentially reducing the impact of dataset's complexity to interpretability problem. The MNIST-equivalent dataset containing pattern theory's elements is designed to facilitate smoother entry into this framework, along which the theory's generative aspect is naturally presented.
\end{abstract}

\section{Introduction}
\label{section:Introduction}
Machine learning and artificial intelligence have taken a leap forward with the success of deep learning (DL), in particular the deep neural networks (DNN). As deep learning gains popularity, its applications have emerged in many different sectors. Several applications require higher degree of accountability, but the black-box nature of a DNN remains a challenge. EXplainable Artificial Intelligence (XAI) trend has thus emerged in response to the challenge of developing more transparent and interpretable algorithms. Various approaches have been proposed (see review XAI papers \cite{8466590, 8400040,Zhang2020ExplainableRA, 9233366}). However, many of them involve post-hoc analysis attempting to provide explanations without clear indications on how they can be utilized to fix or improve the algorithms. Furthermore, different contexts may require different explanations, and thus proper scope of the context may need to be defined rigorously to quantify the correctness of explanations. 

There are few attempts to uncover the fundamental structures within datasets that can be reliably related to the DNN. Consequently, explanations may become equally unstructured and empirical, frequently presented as attribution values that require more refinement. Finding suitable framework to extract and relate datasets' and DNN's fundamental structures is indeed a daunting task, given that there is no guarantee a mathematically well-defined framework to capture their variability exists. Inherent ambiguity w.r.t human perception is another possibly insurmountable hurdle. Nevertheless, related attempts exist \cite{tSNECNN, DBLP:journals/corr/NguyenYC16}, including activation optimization methods \cite{Olah_2017,olah_buildingblock} that yield insights delivered with excellent interactive interfaces. They produce visually intuitive results that could potentially serve as further feedback for the development of more interpretable and manageable algorithms, though we have not discovered such ground-breaking attempts. This paper aims to step closer towards studying XAI through above-mentioned fundamental structures of patterns, adopting the concept of ``analysis by synthesis" \cite{YUILLE2006301} in conjunction with convolutional neural network (CNN), the specific DNN suitable for computer vision. In particular, we revisit general pattern theory (GPT) which studies mathematical objects called \textit{generators} as the basic elements arranged into \textit{configurations}. 

\begin{figure*}[]
  \includegraphics[width=1.0\textwidth , trim = {0 0 0 0cm}]{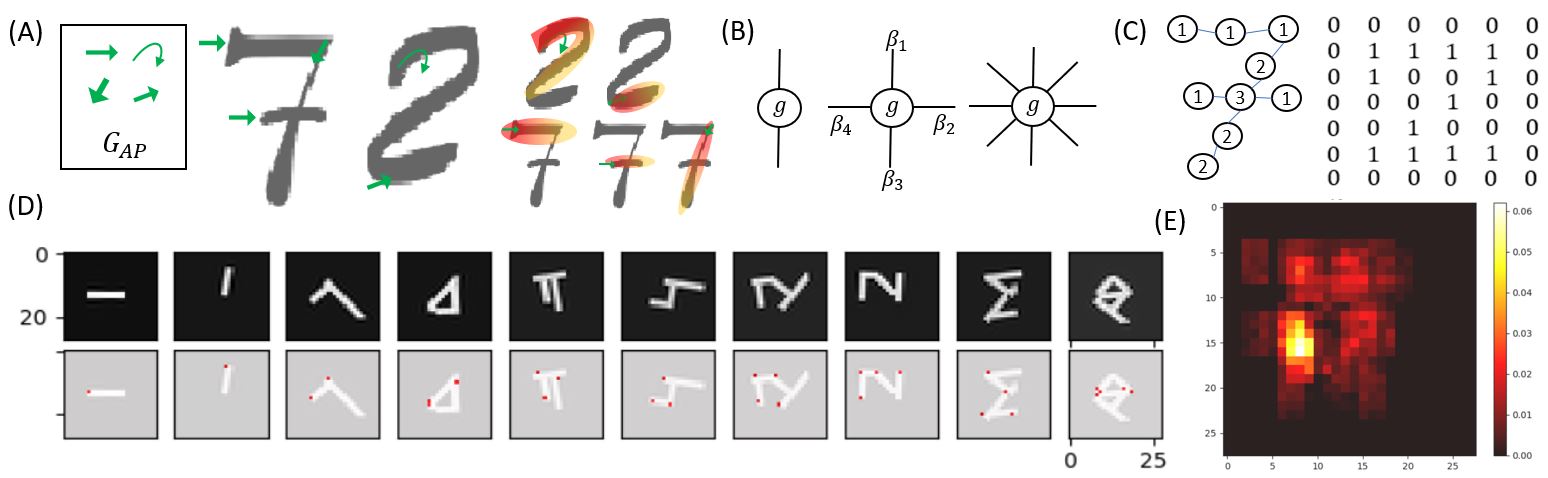}
  \caption{(A) An automated painter with finite \(G_{AP}\). Green arrows denote generators. Read/yellow overlays are idealized heatmaps for component-wise explanations. (B) Generators with arity 2, 4 and 8. Bond values shown are arranged according to a specific topology matrix \(J\) with top/right/bottom/left arrangement. (C) An arbitrary configuration and a square lattice configuration. (D) (top) GPT MNIST images, each representing one of the ten classes. (bottom) The same images with their corresponding generators shown in red dots, shown in increasing number of generators from left to right. (E) GPT MNIST spatial probability distribution of generators, indicating the total chance of a generator being spawned in the \(28\times 28\) square lattice if samples are drawn uniformly from the ten classes. More can be seen in \textit{main supp. material} section \textit{More Heatmaps}.}
  \label{fig:intro}
\end{figure*}

This paper considers CNN that is trained on large datasets using loss minimization, where weights are mainly adjusted by gradient descent or its variants. A popular type of XAI methods on CNN is heatmap-based, but heatmaps may not serve as strong explanations \cite{Tjoa2020QuantifyingEO}. In this paper, we isolate components (generators \(g\in G\)) within each image and extract more targeted explanations based on these generators in the form of local heatmaps. In other words, for each image sample, find a set of attribution values for all its meaningful components \(\{h_g:g\in G\}\). To illustrate this, see fig. \ref{fig:intro}(A). Imagine there exists an automated painter whose ability to paint is constrained to a finite list of strokes such as \(G_{AP}\). The figure shows how 7 and 2 can be created via proper placement of each respective generator \(g\in G_{AP}\), shown as green arrows. Then, ideally, each generator can be assigned an explanation \(h_g\) shown as red-yellowish heat regions. GPT itself is an abstract framework, but a representation of generators and configurations can be seen in, for example, in fig. \ref{fig:intro}(B) and (C).

To obtain these targeted explanations, we construct GPTNet, a DNN architecture that leverages on ResNet's classification power and U-Net's ability to solve difficult semantic segmentation problems. GPTNet performs both classification and the prediction of generators configuration, the latter being a type of semantic segmentation. GPTNet will learn not only features for classification, but its parameters will also be adjusted to distinguish various salient points, which are the locations and transformations of non-zero generators. Existing saliency-based XAI methods will then be applied to this architecture. Section \ref{sect:RW} reviews related works and how we adapt their ideas. Subsection \ref{subsect:GPT} provides minimalist reviews for GPT concepts enough to understand this paper. Section \ref{sect:Expt} describes the experimental setups, from the dataset, GPTNet architecture, training pipeline to the evaluation methods used. Section \ref{sect:results} discusses the results and provides more explanations to figures displayed throughout the paper. Section \ref{sect:conclusion} concludes the paper and provides a brief remark on future developments.

\section{Related Works}
\label{sect:RW}
ResNet \cite{DBLP:journals/corr/HeZRS15} is a well-known DNN architecture with excellent performances in many computer vision tasks. It has many layers and uses skip connections designed to solve vanishing gradient problem. U-Net \cite{DBLP:journals/corr/RonnebergerFB15} excels in semantic segmentation, and has been shown to perform well on ISBI challenge after training on few images. We construct GPTNet, a U-Net-like architecture, but whose encoder consists of ResNet34's parts, to leverage on their strengths in classifications and segmentation-like tasks. Being a deep CNN, both are incompletely understood. This paper performs XAI study and compares a few existing XAI methods to understand how information is propagated along these deep networks.

Generative Adversarial Networks \cite{Goodfellow2014GenerativeAN} has been a milestone in generative modeling. Generative model G is trained along a discriminative model D, optimizing each other such that G generates samples that are less and less distinguishable from the original distribution, while D becomes more adept at identifying the fake samples generated by G. Many derivatives of GAN have come along with astonishing performance \cite{Radford2016UnsupervisedRL, DBLP:journals/corr/abs-1710-10196, Brock2019LargeSG}, including applications such as style transfers and translations \cite{zhu2020unpaired, isola2018imagetoimage, Zou2020StylizedNP} etc, just to mention a few models related to image generation. Since their structures are based on CNN architecture, the mechanisms underlying their success are similarly not completely understood. While conditional GAN does provide a glimpse into a systematic way of generating images from certain classes, random vectors fed into G and their arrangements in the feature spaces are among the complex problems of DNN interpretability. This paper uses GPT framework, where the basic units called generators and their configurations are intended to serve as structured objects for generative modeling tasks. Unlike GAN, generative models under this framework will possess a lower level structure
that can be studied with XAI in a component-wise manner.

XAI methods are developing fast and their applications emerge in different fields, as seen in XAI review papers mentioned earlier. Well-known methods include Local Interpretable Model-Agnostic Explanations (LIME) \cite{10.1145/2939672.2939778}, DeepLIFT \cite{Shrikumar2017LearningIF}, Class Activation Mapping \cite{CAM}, SHapley Additive exPlanations (SHAP) \cite{Lundberg2017AUA} that is based on game-theoretic concept, layerwise relevance propagation (LRP) \cite{PixelWiseLRP} and their many derivatives. This paper specifically compares deconvolution \cite{Zeiler2014VisualizingAU}, its modified version Guided Backpropagation \cite{Springenberg2015StrivingFS} and GradCAM \cite{GradCAM}. 

Various metrics have been devised for the evaluation of XAI methods. This section briefly describes those listed in \textit{related work} section of \cite{Tjoa2020QuantifyingEO}. The quality of heatmaps has been measured by how much the heatmaps improve ILSVRC localization \cite{CAM,GradCAM}. Metrics have been designed to measure changes after the manipulation of pixels with high relevance according to the XAI methods used \cite{PixelWiseLRP,7552539,NIPS2019_9167,Hartley_2021_WACV}, one of which, the \textit{most relevant first} (MoRF) will be adopted here.  We also use weight randomization sanity check in \cite{NIPS2018_8160}, where similarity is measured between explanations given before and after layer weight modification, for example, using \textit{rank correlation}. Such similarity measures are also used in \cite{ancona2018towards,Sixt2020WhenEL}. Both \cite{oramas2018visual} and \cite{Tjoa2020QuantifyingEO} design synthetic datasets with ground-truth heatmaps. The former measures IoU between generated heatmaps and ground-truth, while the latter computes precision, recall and ROC. Unlike them, this paper uses \textit{configurations} rather than feature/localization masks as ground-truths, though configurations appear like very sparse masks. The evaluation procedure called the \textit{pointing game} \cite{8237633,9157775} is unlikely to be meaningful here, since comparing hits/misses between a maximum value in non-sparse heatmaps with sparse configurations will come with many subtleties. Instead, simple pixel-wise hits/misses like recall and precision are more suitable to quantify explanations w.r.t configurations because class imbalance will be accounted for.

\subsection{General Pattern Theory}
\label{subsect:GPT}
We closely follow the textbook by Ulf Grenander himself \cite{grenander1993general}, the mathematician formulating a significant portion of the theory. While the theory has evolved to include a lot of aspects related to statistics and applied mathematics, we focus on its GPT framework. As a start, we review GPT basic definitions, only briefly going through concepts not directly relevant here, and then immediately adopt the ideas presented as APL codes in the textbook's \textit{abstract biological patterns} chapter. The github link to the project is \url{https://github.com/etjoa003/gpt}, which includes python translation of the APL codes. Also see a summary note on pattern theory \cite{ptnote}. 

In this paper, we are interested in images as the realizations of objects in GPT. Define the \textit{initial generator space} \(G_0\) and \textit{bond structure group (BSG)}. BSG is an example of the implementation of a \textit{similarity group} \(S\) (details not relevant here). A generator \(g\in G_0\) is also denoted by \(\alpha\) and has \(\omega (g)\) bonds attached to it. \(\omega(g)\) is called the \textit{arity} of \(g\). Also, \(g\) is abstract; it can be defined as \(g\in\mathbb{Z}\), \(g\in\mathbb{R}^3\) etc, depending on the system. Each bond has a \textit{bond value} \(\beta_j(g)\in B_s(g)\) for \(j=1,\cdots,\omega(g)\). BSG is the permutation group of \(B_s(g)\). Then the \textit{generator space} \(G=\{sg:\forall g\in G_0, \forall s\in BSG\}\). Example 1: fig. \ref{fig:intro}(B) shows \(\omega(g)=2,4,8\). Example 2: \(BSG=\{s_1,s_2,s_3,s_4\}\) is the cyclic permutations of the bond values for \(\omega(g)=4\) e.g. \(s_1 g=g\), also \( \beta_j(g)=\beta_{j'}(s_2 g)\) where \(j'= 1 + (j\ mod\ 4)\), \(s_3 s_3=s_1\) etc. In this paper, the generator space will be a partition \(G=\cup_\alpha G^\alpha\), where \(G_\alpha=\{s\alpha:\forall s\in BSG\}\), i.e. BSG elements do not transform one generator to another. 

\begin{figure}[h!]
\centering
  \includegraphics[width=0.32\textwidth , trim = {0 0 0 1.1cm}]{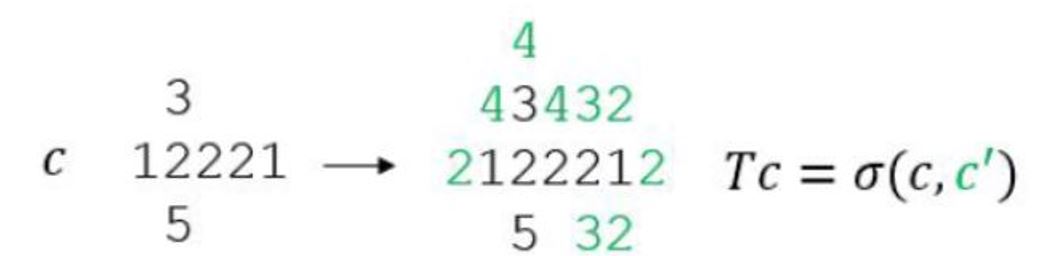}
  \caption{An example of configuration transformation \(T\) in a lattice square. Let \(G=\{0,1,\dots,5\}\), arity \(4\), \(\beta_i(g)=g\) for \(i=1,2,3,4\). Given configuration \(c\), a new generator \(g\) is placed in a \(Tc\) site if it is equal to the maximum value of all its neighbours, unless the maximum value is \(r\). Here, \(r=5\) and \(0\) is not shown.}
  \label{fig:Texample}
\end{figure}

A \textit{configuration} \(c=\sigma(g)=\sigma(g_1,\dots,g_n)\), where \(\sigma\in\Sigma\), also called the \textit{connector}, is a graph with \(n\) sites and segments\footnote{In modern graph theory, sites and segments are respectively vertices and edges.}. At each site \(k\), a generator \(g_k\in G\) is placed. Each segment links two generators depending on a given rule typically set by the specification of bond relation \(\rho\). Define for any \(g,g'\in G\) the \textit{bond relation} \(\rho\) such that \(\rho_{jj'}=\rho(\beta_j(g),\beta_{j'}(g'))\) is valid if and only if \(\rho_{jj'}=1\) and zero otherwise. This is the simple deterministic case, though it can be probabilistic too. The \textit{configuration space} is \(\mathcal{C}=(G,S,\Sigma)\). If \(\rho\) is defined, \(\mathcal{R}=(G,S,\rho,\Sigma)\) is the \textit{regularity}. \textit{Regular configuration} is denoted \(\mathcal{C}(\mathcal{R})\). Here, we use \(\rho\) without explicit definition, as the legitimate bond relations can be implicitly set through the definition of \textit{configuration transformation} \(T:\mathcal{C}\times \mathbb{Z}_+ \rightarrow\mathcal{C}\) relevant to our experiment, where \(\mathbb{Z}_+=\{1,2,\dots\}\) is the time-step; example shown in fig. \ref{fig:Texample}. Recursive notation can be used naturally \(Tc_{k+1}=c_k\) or when no ambiguity arises. The concept of \(T\) is introduced in chapter 4 of \citep{grenander1993general} to describe \textit{abstract biological patterns}. It helps to think of each generator as a cell, and the transformation as the growth of cells colony; thus a transformation can be suitably called \textit{growth function}. The design of growth functions and the choices of generator space have been demonstrated to produce a rich variety of patterns. 

Many topics in Pattern Theory (PT) have been compiled \cite{grenander1993general, Grenander2007PatternTF}. PT framework has also been used, for example, to study semantic structures in videos \cite{DESOUZA201641} and biological growth \cite{4162636}. However, we believe the use of GPT for XAI study of CNN has not been previously done. With modern computing capability, in the near future, the integration of PT framework and machine learning models may give rise to models that are more interpretable via the analysis of structures like those described above, while leveraging on DNN's success as predicitive, semantic segmentation or generative models. This paper particularly demonstrates how heatmap-based XAI methods can yield component-wise saliency maps for configurations' individual part uncovered by our DNN, the GPTNet.

\begin{figure}[htpb]
  \includegraphics[width=0.5\textwidth , trim = {0 0 0 0}]{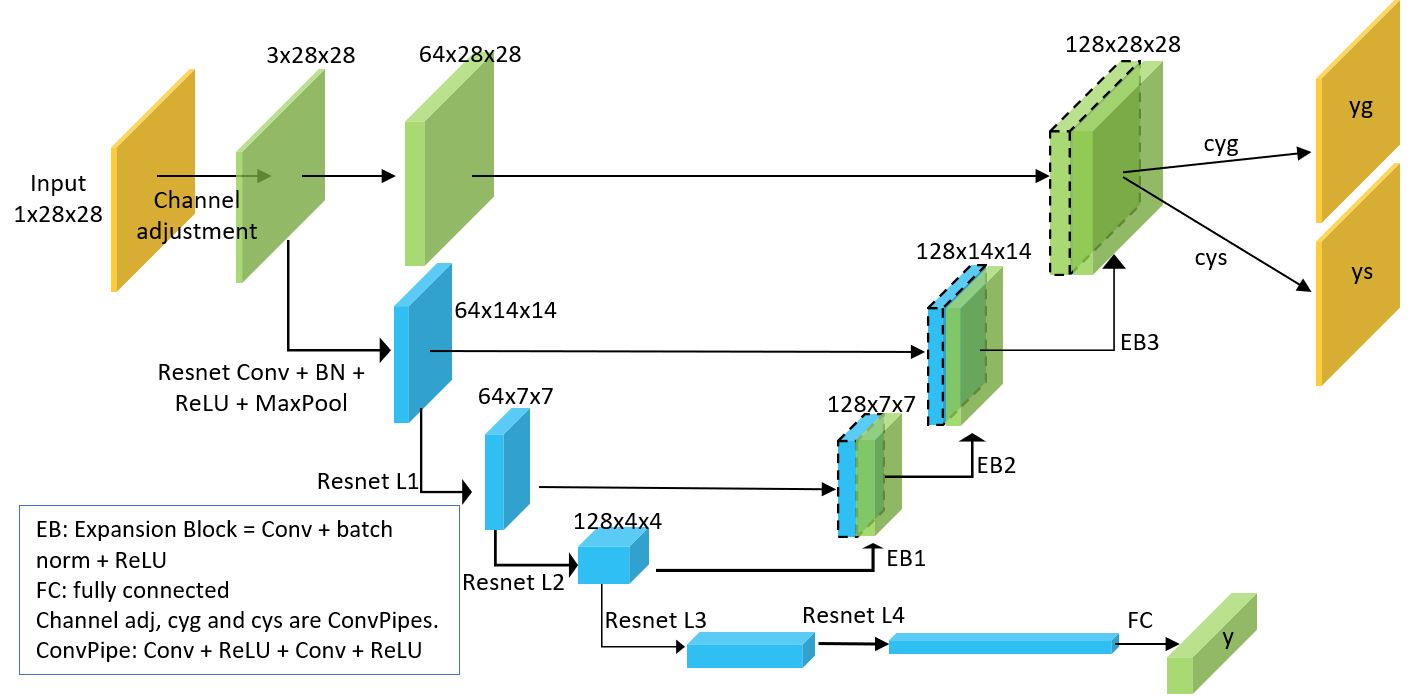}
  \caption{GPTNet architecture, combining components of ResNet34 (shown in blue) with convolutional blocks to form U-Net-like architecture. Details of ResNet components are not shown. Output of FC, cyg, cys are \(y\) for class predictions, \(y_g\) for ``semantic segmentation" of generator types and \(y_s\) of generator transformations respectively.}
  \label{fig:arch}
\end{figure}

\begin{figure*}[t]
  \includegraphics[width=1\textwidth , trim = {0 0 0 0}]{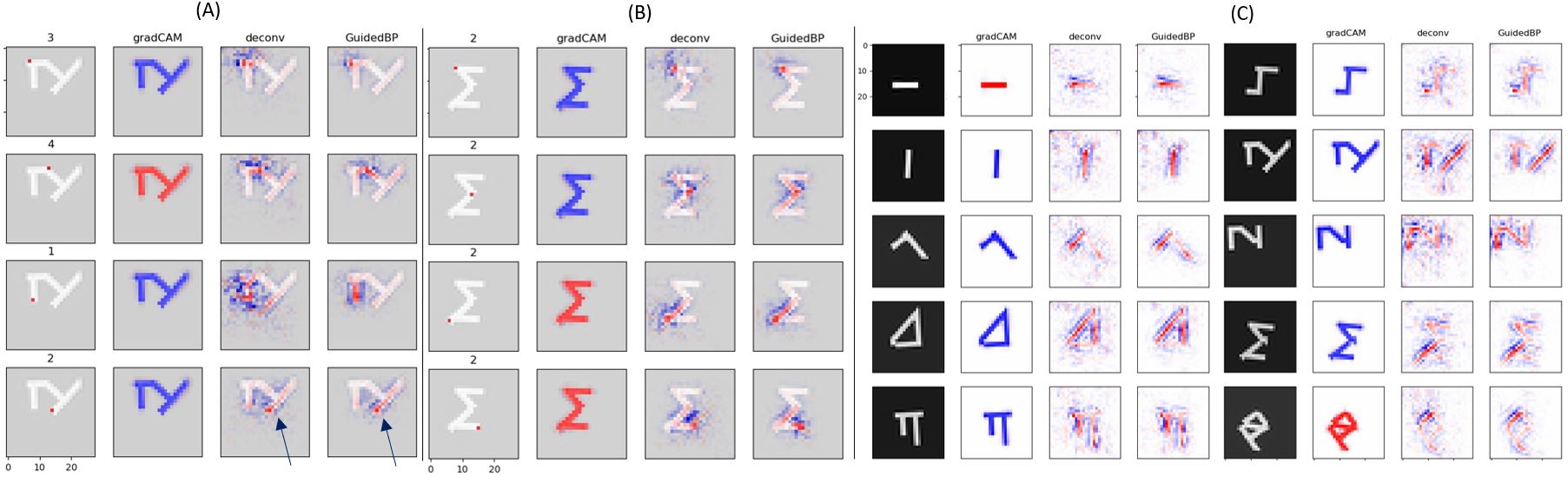}
  \caption{(A) (column 1) A sample \(y_s\) output for an image of class \(c=6\), showing predicted generators position (red dots) and their corresponding \(y_s\) label. Different rows correspond to different generators from the same image. (columns 2 to 4) Heatmaps for each generator are generated using XAI methods gradCAM, deconv and GuidedBP. Heatmaps demonstrate the potential for component-wise, generator-specific explanations, some with directions marked correctly (see arrows). (B) Same as (A) but for \(y_g\) on a \(c=8\) sample. (C) Standard heatmaps from class prediction \(y\) without component-wise explanation, one for each class. Remark: for all heatmaps, the maximum and minimum values displayed are 1 (red) and -1 (blue) respectively, matching the normalization performed during post-processing. Zero value in the heatmaps is shown as white. Grey background in (A) and (B) are due to white region in the heatmaps shown with some level of transparency over the black background of GPT MNIST digits.}
  \label{fig:heatmaps}
\end{figure*}

\section{Experiments}
\label{sect:Expt}
The dataset used here, the GPT MNIST, will be a MNIST equivalent for studying underlying structure of images under GPT framework. Ten different patterns like MNIST digits can be synthesized and sampled on demand using a sampler class available in the code repository, as shown in fig. \ref{fig:intro}(D) (see more in the \textit{dataset} section of \textit{main supp. material}). The connector \(\sigma\in\Sigma\) is a \(28\times 28\) square lattice, corresponding to \(28\times 28\) pixels. All generators have \(\omega=8\), each forming a square with 8 directly adjacent neighbors, also known as the Moore neighborhood. The topology is defined as matrix \(J\) (see below) so that the first row indicates the neighbor on top and the following rows running clockwise. \(G\) is generated from \(G_0=\{\alpha:\alpha=0,1,2,3\}\) (represented by matrix \(G0\) below) and cyclic permutation \(BSG=\{s_i:i=1,2,\dots ,8\}\). Each row of \(G0\) corresponds to a generator. The entries in \(G0\) are bond values. With the definition of growth function used here (described in the next paragraph), the value \(G0_{ij}=\beta_j(i)\) corresponds to the placement of a new generator identified by \(\beta_j(i)\)-th row of \(G0\) at the neighbouring site in \(j\)-th bond direction, and transformed by \(s_j\), assuming \(i=1,2,3,4\) and placement is valid. The value 0 in \(G0\) means no bond. Row 4 (or \(\alpha=3\)) is a terminal generator, roughly preventing any further growth in its direction. Row 2 (\(\alpha=1\)) will spawn a copy of itself, a \(\alpha =2\) that lines along the newly spawned \(\alpha =1\), and a terminal generator \(\alpha =3\), leading to the formation of a straight stroke. \(\alpha =0\) and \(3\) have the same behavior, but the latter is considered a ``living" cell and not empty. Do see  \textit{indexing subtleties} in the \textit{main supp. material}.

\[
J=
\begin{psmallmatrix}
    0  &  1  \\
    1  &  1 \\
    1 & 0 \\
     1 & -1 \\
     0  & -1\\ 
     -1 &-1\\
     -1 & 0 \\
     -1 & 1
\end{psmallmatrix}
,G0= 
\begin{bmatrix}
    0&0&0&0&0&0&0&0      \\
    2&3&4&0&0&0&0&0      \\
    0&0&0&0&0&0&0&2      \\
    0&0&0&0&0&0&0&0      
\end{bmatrix} 
\]

Each image in fig. \ref{fig:intro}(D) is generated using \textit{DEVELOP} function applied on initial configuration \(c_0\), where \(c_0\) is a lattice with empty generators \(\alpha=0\) everywhere in the lattice except the red dots (orientation not shown). Each generator is technically \(sg\in G\) with \(s\in BSG, g\in G_0\). Thus, a configuration \(c\) can be represented by both array \(y_g\) whose entries correspond to \(g\in G_0\), and array \(y_s\) whose entries correspond to \(s\in BSG\) (orientation). The implementation of \textit{DEVELOP} consists of modified \textit{GROWTH1} function and \textit{COMPENV} from the textbook. GROWTH1 is a configuration transformation, specifying the \textit{dynamic rule} that evolves the configuration according to the generator settings (\(G,BSG\) etc) and the \textit{environment} computed by COMPENV. In this experiment, each generator produces a stroke (generators are spawned along a straight-line) that, roughly speaking, stops growing when it hits other generators. The original growth function uses the \textit{dynamic rule} that adds a new generator on each empty site \textit{with only one non-empty generator neighbor contributing non-zero bond value} \(\alpha+1\) there (the spawned generator is then identified by \(\alpha\), corresponding to the \(\alpha\) notation used so far). The modification here allows for generator to be placed at an empty site even if there are more than one non-empty generator neighbors contributing non-zero bond values. With \(G0\) we have defined, this helps prevents patterns with holes. More explanatory details are in \textit{main supp. material} section \textit{DEVELOP function}.

The architecture used here is GPTNet, shown in fig. \ref{fig:arch}. Components from pre-trained ResNet34 is loaded into GPTNet, and then expansion blocks (EB) upsample feature maps starting from ResNet L2 layer. The input size is \(1\times 28\times 28\), a 1 channel image like MNIST. The outputs are \(y,y_g,y_s\), where \(y\) corresponds to class prediction. As usual, \(y\) has 10 channels, each channel \(y_c\) for a class, and \(c_{pred}=\displaystyle arg\max_{c} y_c\) is the predicted class. \(y_g,y_s\) are the outputs for predictions on configuration and its transformations, each pixel having 4 and 8 channels, respectively corresponding to the number of generators in \(G0\) and the number of transformations in \(BSG\). Similar to \(y\), argmax is used for pixel-wise prediction.  

The training process includes a training stage (TS) and two fine-tuning stages (FT1 and FT2), all using adam optimizer and cross-entropy losses. In TS and FT1, samples are drawn uniformly from all classes. Batch size 16 is used. Other settings vary between the phases. In TS, adam optimizer is used with initial learning rate \(lr=10^{-3}\), weight decay \(10^{-5}\) and \(\beta=(0.5,0.999)\), though when soft targets are reached for all three cross-entropy losses of \(y,y_g,y_s\), learning rate is reduced to \(lr=10^{-4}\). Training is terminated early when all three target losses are reached. During FT1, similar settings are used for adam optimizer except \(lr=4\times 10^{-5}\), and fine-tuning proceeded for 10 epochs of 12800 iterations without achieving target losses we specified. Class imbalance is present in \(y_g,y_s\) mainly due to the empty generators and the use of upwards as the default orientation. Hence, less weights are assigned to their losses in TS to prevent mode collapses. This is done by setting \(0.01\) to the loss weights on the channels corresponding to these classes and setting \(1\) to the rest. In FT1, all losses are treated equally, resulting in convergence for the \(y_g,y_s\) of most classes \(c=0,1,\cdots,7\). FT2 is a comparatively short process to fine-tune the convergence of the more difficult classes \(c=8, 9\) with human-in-the-loop. Hyper-parameters are adjusted and FT2 repeated several times until seemingly the most convergent parameters are found. This includes the adjustment of sample distributions to draw more instances of the difficult classes. From visual inspection of the images generated from models after FT2, once in a while there is a component that is wrongly oriented. Regardless, the final model chosen for evaluation is the one with best loss performances obtained during FT2. For all training details, see the \textit{main supp. material}, section \textit{training}.

After model training, gradCAM, deconvolution and Guided Backpropagation (BP) are used to generate heatmaps to explain the particular prediction, as shown in fig. \ref{fig:heatmaps}. The figure shows (A) component-wise saliency maps derived from each target pixel \(y_s[i,j]\) for a \(c=6\) sample where \((i,j)\) are the coordinates where non-zero generators are predicted to be, (B) likewise component-wise saliency from \(y_g[i,j]\) for a \(c=8\) sample, and (C) heatmaps derived from \(y\) one for each class. Basic metrics and Area over Perturbation Curve (AOPC) are computed, along with Spearman's Rank Correlations (SRC) for cascading weight randomization. Results are discussed in the next section.

\textit{Cascading weight randomization} is performed by first computing SRC for each data sample. Let \(F\) denotes the GPTNet, \((y,y_g,y_s)=F(x)\), where \(y\in\mathbb{R}^{10}\) is the class probability vector; as before, \(c_{pred}=\displaystyle arg\max_{c} y_c\). For each individual sample,  let \(h_{xai}(y,F)\) be the processed heatmaps where xai = gradCAM, deconv or GuidedBP. Processing includes normalization over absolute max values. We also plot SRC for the so-called \textit{diverging visualization} (no ABS) and \textit{absolute value visualization} (ABS) \cite{NIPS2018_8160}, the latter applying absolute value function to all attribution pixels as a part of the processing procedure. Furthermore, when too many layers are perturbed, we do observe all zero attributions, which we then process by replacing them with small random values. For class prediction, first, ResNet's fully-connected (FC) layer weights are randomized, yielding the perturbed vector \((y^{(1)},y_g^{(1)},y^{(1)}_s)=F^{(1)}(x)\). Then a sample SRC is computed by \(SRC[h_{xai}(y, F),h_{xai}(y^{(1)},F^{(1)})]\). Next, repeat the process by perturbing ResNet L4 layer of \(F^{(1)}\) to get \(F^{(2)}\), then L3 layer from \(F^{(2)}\) to get \(F^{(3)}\) and so on, SRC still computed against the unrandomized version \(SRC[h_{xai}(y, F),\ .\ ]\). Likewise, for \(y_g\), first, layer \(cgy\) is perturbed, then expansion block 2 (EB2) and so on; likewise \(y_s\). SRC values discussed in result section are averages over data samples.

MoRF evaluation is performed by computing AOPC. Let \(O=(r_1,\dots,r_L)\) be an ordering such that \(r_1\) is the most relevant pixel (highest attribution) and \(r_{L}\) the least relevant of the top L most relevant pixels. Define generally \(g\) such that \(g(x,r_k)\) removes some pixels around and including \(r_k\). In this paper, due to small lattice size, \(g\) removes only \(r_k\) and replaces it with a random value. Let \(\Delta^k=f(x_{MoRF}^{(0)})-f(x_{MoRF}^{(k)})\) be the gap between original and perturbed confidence value, \(\langle .\rangle\) be average over data distribution, the recursive form of the perturbation be \(x^{(k)}_{MoRF}=g(x^{(k-1)}_{MoRF},r_k)\) and \(x^{(0)}_{MoRF}=x\) and \(f(.)=F_c(.)\) be the channel for class \(c\) prediction. Then, under MoRF framework, as defined in \cite{7552539}, \(AOPC=\displaystyle\frac{1}{L+1}\big\langle \sum_{k=0}^{L}  \Delta^k \big\rangle\).

\begin{figure*}[t]
  \includegraphics[width=1\textwidth , trim = {0 0 0 0}]{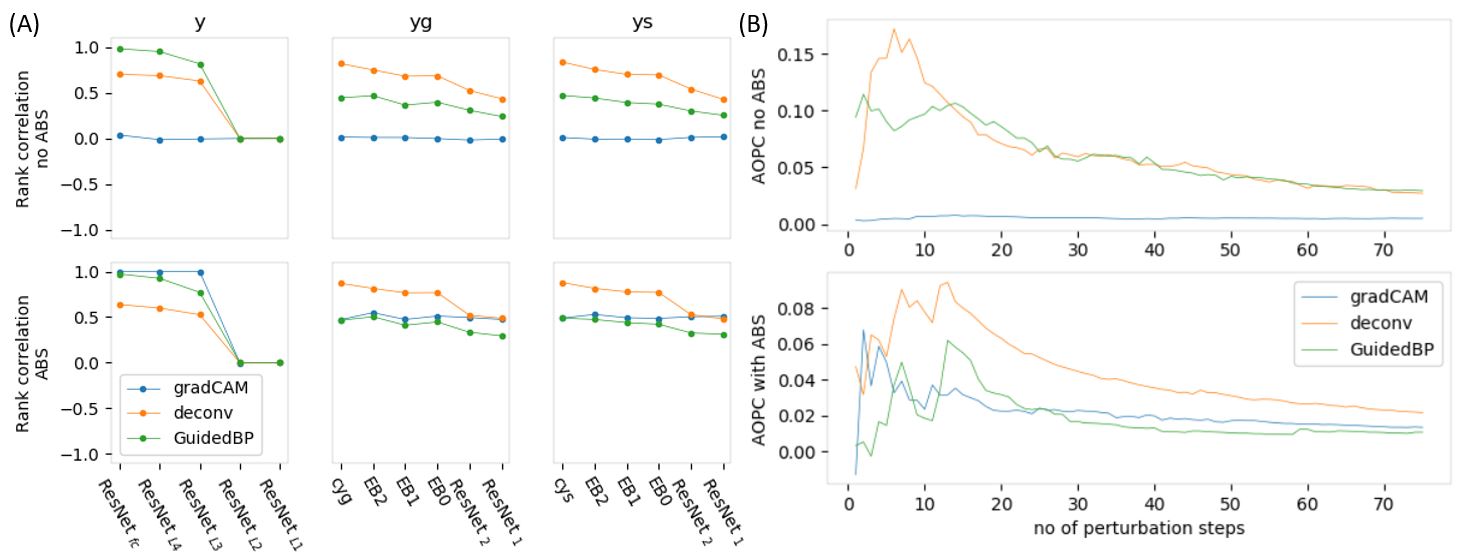}
  \caption{(A) Plots of Spearman's rank correlations for the different output channels. Within each graph, cascading weight randomization proceeds from left to right. (B) Plots of Area over Perturbation Curve (AOPC) for \(y\) output channel.}
  \label{fig:metrics}
\end{figure*}

\section{Results and Discussions}
\label{sect:results}
Accuracy, precision and recall are evaluated for 2400 test samples generated adhoc. GPTNet attains error rates of \(<0.02\%\) except for \(y_{s}\) precision as shown in table \ref{tab:acc}. Using the combination of powerful deep learning architectures, excellent performances of class prediction and semantic segmentation-like process on this relatively simple data are easy to attain.

\begin{table}[htpb]
\caption{Hit/miss evaluation metrics by \(\%-\)error on 2400 test samples, averaged over pixels when applicable. Classification accuracy (not shown) is \(100\%\).}
\begin{center}
\begin{tabular}{cccc}
    \hline
    &acc.&recall&precision\\    
    \hline
    \(y_g\)&\(<0.01\)&\(0.\)&\(0.0104\)\\
    \hline
    \(y_s\)&\(<0.01\)&\(0.\)&\(0.254\)\\
    \hline
\end{tabular}
\end{center}
\label{tab:acc}
\end{table}

The core idea in this paper revolves around the concept demonstrated by the automated painter described in the introduction. If we can uncover the underlying generators distribution realized as variables \(y_g\) (and their transformations as \(y_s\)), then, for an image, a set of explanations can be given as the saliency maps covering the corresponding components in the image. From a given prediction, we can then extract explanations whose complexity of interpretability is upper-bounded by the knowledge we have over the set of our constraints (\(G,BSG\), growth functions etc). Fig. \ref{fig:heatmaps}(A) illustrates this through the decomposition of the images into a series of generator transformations positioned at the red dots; the knowledge we specifically acquire from them is the directions of the strokes. The numbers above the images denote the type of BSG applied, e.g. 1 means default orientation, 3 means turned clockwise twice, thus the generator faces \(90^\circ\). Fig. \ref{fig:heatmaps}(B) similarly shows the generators located at the red dots, the numbers indicating the type of generators (number 2 for row 2 of \(G_0\) or equivalently \(\alpha=1\); indeed they are the only generators in the initial configurations used to generate the strokes). In both cases, gradCAM appears incapable of providing any discriminating information. Guided BP and deconvolution appear to have good potentials in indicating the true areas covered by the rotated generators. The last two rows of fig \ref{fig:heatmaps}(A) show strong solid responses in their respective directions (1 for upwards, 2 for up-right). Likewise, the generator transformed with 4 shows strong positive regions pointing bottom-right (recall rotation is clockwise on Moor neighborhood). By contrast, fig. \ref{fig:heatmaps}(C) shows heatmaps for \(y\) prediction one for each class, where the individual components are not separated. We obtain regular heatmap outputs shown before by many XAI research papers showing heat regions that are sometimes unclear, which implies that that L3 and L4 layers do remove potentially useful information for obtaining explanations.

Heatmaps generated by existing methods lack the power of explanations when they simply highlight areas that correspond to localization. Distinguishing a dog from a cat can be, for example, more clearly justified by providing a component of evaluation that uniquely identify snouts. However, existing methods may instead relegate its explanations to just highlighting these features, losing localization information or possibly other important features, such as the background context. This paper addresses these problems. By presenting a series of relevant components like fig. \ref{fig:heatmaps}(A,B), (1) human readers can synthesize specific information about a data sample from the individual components or from the semantic sum of all components (2) we can also understand more about DNN from inspecting different meaningful components it delineates and focuses on. The notion of using gestalt laws and spatial grouping rules for ``analysis by synthesis" in \cite{YUILLE2006301} have therefore been extended here to the components of heatmaps, each of which closely corresponds to  the \textit{generator}, i.e. the synthesis component.

SRC in fig. \ref{fig:metrics}(A) can be read in the following manner: as cascading weight randomization progresses, SRC of better heatmaps drop/increase to zero faster from positive/negative. For heatmaps generated on \(y\) output, GradCAM with no ABS show zero SRCs. This is because attribution values' signs clearly affect the SRC computations. Its heatmaps appear to show attributions with all red/positive or all blue/negative along GPT MNIST digits, verified by visual inspection (see fig. \ref{fig:heatmaps}(A,B) etc). Consequently, they average out SRC to zero. However, as indicated by its ABS plot, the first three randomization appear to yield no effect, confirming the results in \cite{NIPS2018_8160}. Guided BP shows weaker response than deconvolution. For example, FC layer randomization almost makes no difference to the attribution i.e. SRC near 1. When cascading randomization progresses, its SRC drops, but remains above deconvolution's.

For \(y_s\) and \(y_g\), GradCAM's extreme polarity renders no-ABS SRC uninformative. For SRC with ABS, on the other hand Guided BP shows stronger response than GradCAM, which is in turn stronger than deconvolution. Interestingly, the constantness of this gradCAM SRC seems to indicate that only cyg and cys layers affect the quality of attributions meaningfully with 0.5 drop. The drop of SRC to a value above zero indicates that the effect of GPTNet's weights do not fully affect the heatmaps. The remaining similarity must thus be the artifact of the inputs; the higher SRC is above zero after late stage randomization, the less effective is the XAI method in filtering the input to extract meaningful explanations. Note that when cascading progresses to late stages, for deep architecture such as ResNet, numerical singularity does occur. We replace them with small random numbers, yielding zero SRCs. 

AOPC plotted in fig. \ref{fig:metrics}(B) shows a trend not previously seen. Small number of perturbations remove the most salient pixels, causing large gaps between original prediction and the perturbed version, i.e. large \(\Delta^k\) at smaller \(k\), hence increasing AOPC. The higher the increase, the more effective is the XAI method. As more salient pixels are removed, the combined effect results in increasing AOPC trend up to 10 to 20 perturbations, except for extreme polar case such as no ABS AOPC for gradCAM. The dip from a peak in AOPC values are not observed in \cite{7552539}. However, there is an intuitive explanation. As more pixels are perturbed, the less important pixels start to be perturbed. These pixels contribute to smaller gaps, i.e. \(\Delta^k\) at larger \(k\) should be small. However, from AOPC definition, this \(\Delta^k\) value is equally divided by \(L+1\), causing the gaps in larger \(k\) to dilute the gaps due to smaller \(k\). It can still be said that deconvolution is more responsive on \(y\) output than Guided BP by observing the first few perturbations. This agrees with cascading randomization results. AOPC is not compatible for \(y_s,y_g\) measurements, as pixel-wise gap w.r.t unperturbed pixel depends on the ground-truth's generators' position apart from labels. For now, there is no sensible way to compare between predicted generators spawned nearby, but not exactly at the correct pixel. In this sense, the GPT MNIST is not a simple dataset due to the precision in spatial position required to produce correct patterns. This constraint can be relaxed in larger lattices and future works using more sophisticated generators and growth functions.

\begin{figure}[h!]
\centering
  \includegraphics[width=0.38\textwidth , trim = {0cm 0 0 0cm}]{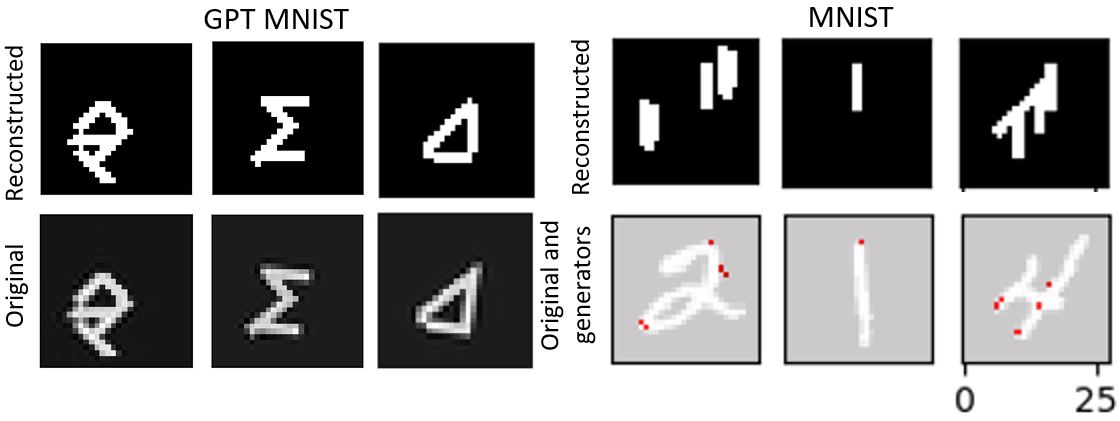}
  \caption{(Left) Reconstruction of images using generators extracted from the original images, showing good reconstruction. (Right) Reconstruction of images using generators extracted from MNIST data, which are not samples from GPT MNIST, yielding poor reconstruction.}
  \label{fig:generative}
\end{figure}

Collecting some of the observations, we see that heuristics such as ``set negative activation to zero" may not always yield better interpretability performance. Guided BP, being technically deconvolution with additional \textit{only positive activation terms} \(f_i^l>0\), appears to score worse than deconvolution in \(y\) explanations but better in \(y_g,y_s\) when measured using SRC. Their heatmaps in fig. \ref{fig:heatmaps} show similarity, although Guided BP appears less noisy. Component-wise explanations are exhibited by both of them, while gradCAM appears to fall short in many aspects. The generative aspect from this model is shown in fig \ref{fig:generative}. For data sampled from the same distribution as the training dataset (see fig. \ref{fig:intro}(E)), reconstruction is excellent. However, when the same model is used on the original MNIST, we do not obtain good reconstruction. This is natural, as no adjustment such as transfer learning to different generator distribution of MNIST has been done. Furthermore, this paper serves as an instructive revisit to GPT concept, and thus the growth function is varied with only few parameters; indeed, most of the growth steps are capped at 12 time-steps, i.e. a generator at most extends to 12 pixels. With limited \(G0\), \(BSG\) and simple growth rule defined for demonstration, we do not expect the system to generalize yet.

\section{Conclusion and Further Developments}
\label{sect:conclusion}
We have demonstrated a different XAI study on a hybrid CNN architecture under General Pattern Theory framework. The main findings include component-wise heatmaps, where components' behavior is constrained by the design of the system which we have knowledge about. Having semantic grasps on components and sub-structures of the system helps us understand more about the whole system, including the data samples and the DNN architecture. By adopting ``analysis by synthesis" concept, we thus possess greater understanding of the whole explanation by combining our knowledge about each generator's role within the configuration in the generative process. We have also compared several XAI methods under this framework and performed sanity checks.

Further developments can be considered mainly along XAI direction and generative modeling. For XAI, GPTNet-like structures may be encouraged in different tasks, where EB modules can be attached to the original architecture for extracting individual generators as component-wise explanations. Solutions may be required to improve training convergence, since EB blocks attachments increase the complexity of DNN architecture. Nevertheless, we hypothesize that they may have resulted in more meaningful parameter-adjustments, leading to better tuning of DNN parameters. For generative models, different definitions of \(G_0, BSG\) and freely parameterized growth functions etc could lead to the generation of more complex patterns. The obvious next step in the generative aspect is to achieve the generative power of GAN for natural images. A well-trained model will be able to extract generator configurations corresponding to natural images, and the corresponding growth function can reverse the process by evolving the extracted configuration into the original images. With parameterized generators, rich variations of generated images related to a particular configuration may even be possible, for example a face seen from slightly different angles. The possibly arbitrary choices of generators and other settings might become a fertile ground for XAI study, where meanings and explanations can be progressively refined.

\section*{Acknowledgements}
This research was supported by Alibaba Group Holding Limited, DAMO Academy, Health-AI division under Alibaba-NTU Talent Program. The program is the collaboration between Alibaba and Nanyang Technological University, Singapore, hosted by Alibaba-NTU Joint Research Institute.

\bibliography{gptmnist}
\bibliographystyle{icml2021}

\onecolumn

\icmltitle{Supplementary Materials}

\setcounter{section}{0}
\section{Introduction to Code Repository}
This is the main supplementary material for \textit{Convolutional Neural Network Interpretability with General Pattern Theory}. We use the concepts in the \textit{textbook} \cite{grenander1993general}. For the main text, chapter 1 and 4 are sufficient. 

The textbook's computer codes are in APL language. We provide the python code version, which we will refer to as the \textit{repo}. The relevant repo version is v0.4. The actual github link is in the main text.

The root folder is called gpt. It contains the following essential folders and files:
\parskip0em 
\begin{enumerate}
\itemsep-0.4em 
\item README.md.
\item src
\begin{enumerate}
\item utils
\item gpt\textunderscore mnist; this is the folder containing codes used in the main paper.
\item APL\textunderscore to\textunderscore python; this folder contains source codes that reproduce the codes in section 4.2 of the textbook.
\end{enumerate}
\item main\textunderscore gpt\textunderscore mnist.py; input arguments for training GPTNet models etc here.
\item main\textunderscore gpt\textunderscore mnist\textunderscore eval.py; for GPTNet evaluations.
\end{enumerate}
\parskip1em 

Other files and folders include \textit{notebooks}, \textit{\textunderscore references}, while folders such as \textit{checkpoint} will be generated when the code is run. 

The pretrained model used in the main text can be found in the footnote\footnote{\url{https://drive.google.com/drive/folders/1ExsjTa_yJ0oko-VTCgnfgiKgNYe9cHgh?usp=sharing}}.

\section{Dataset}
\begin{figure*}[h!]
\centering
  \includegraphics[width=1\textwidth , trim = {0 0 0 0}]{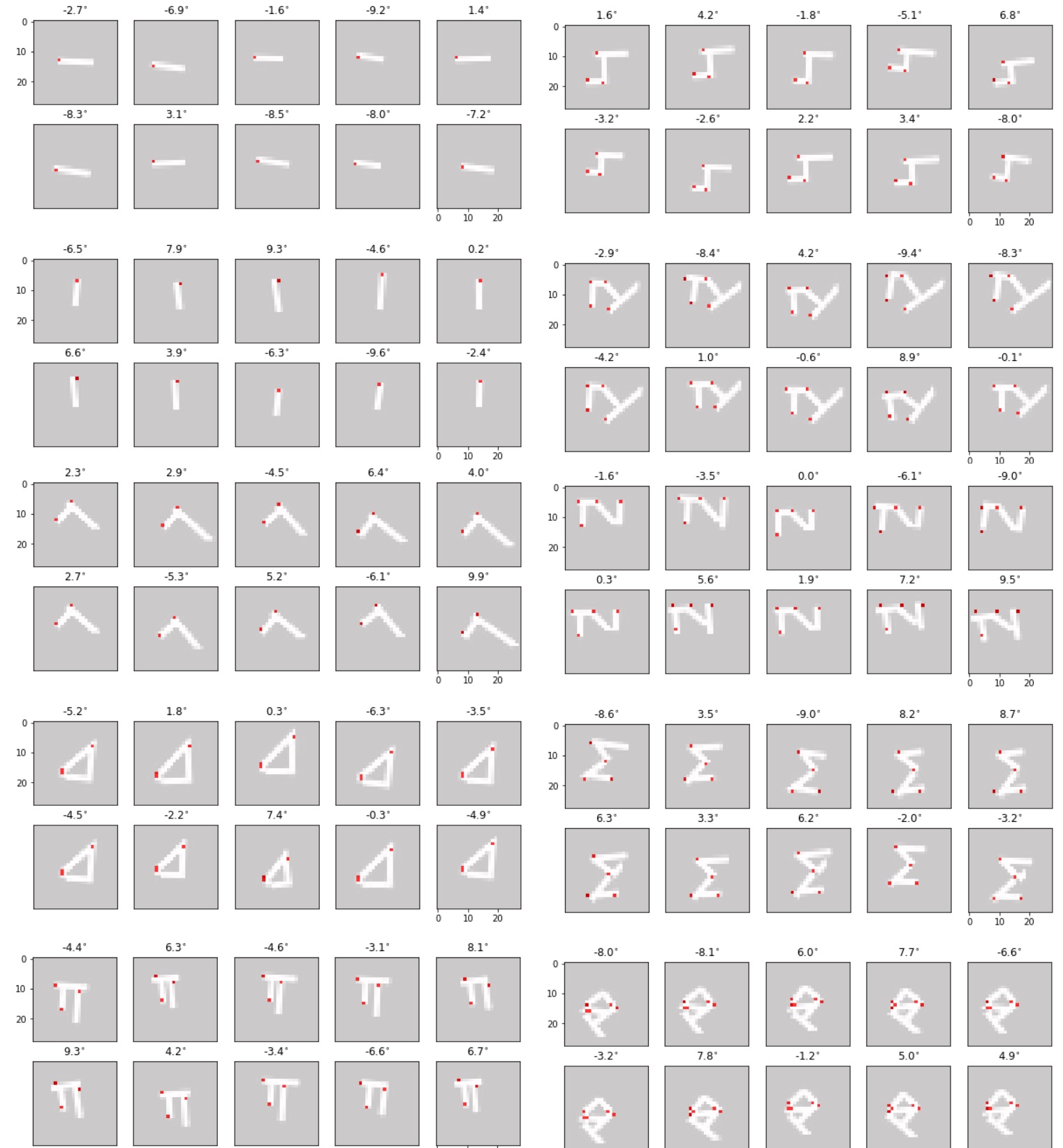}
  \caption{Showing variation of data within each class. Transformation by \(BSG\) elements not shown.}
  \label{fig:dataset}
\end{figure*}

The dataset used is called GPT MNIST, shown in fig. \ref{fig:dataset}. The dataset consists of synthetic data which are synthesized on demand. Each batch of data can sampled by using the sampler code found in gpt\textunderscore mnist/sampler.py, as demonstrated in notebooks/tutorial\textunderscore gpt\textunderscore mnist\textunderscore data.ipynb. The sample generation procedure is as the following: choose a class from \(c=0,1,\dots,9\). Given a class \(c\), choose a random set of positions for generators which are centered around \(n_c\) positions that are specific to the class, i.e. \(P_c=\{(x_i,y_i):i=1,\dots,n_c\}\). Fixed types of transformations \(s\in BSG\) are applied on generators according to their centers \(P_c\), forming consistent recognizable shapes. Use growth function mentioned in the main text to grow these generators. In this experiment, the growth function for each class takes from 8 to 12 time-steps as the argument (randomly sampled). Finally, small rotations are applied onto the images.

\section{GPTNet Architecture}
\parskip0em 
Pre-trained ResNet34 model from torchvision consists of the following layers, of which we use 0 to 7:
\begin{itemize}
\itemsep0em 
\item 0. class 'torch.nn.modules.conv.Conv2d'
\item 1. class 'torch.nn.modules.batchnorm.BatchNorm2d'
\item 2. class 'torch.nn.modules.activation.ReLU'
\item 3. class 'torch.nn.modules.pooling.MaxPool2d'
\item 4. class 'torch.nn.modules.container.Sequential'
\item 5. class 'torch.nn.modules.container.Sequential'
\item 6. class 'torch.nn.modules.container.Sequential'
\item 7. class 'torch.nn.modules.container.Sequential'
\item 8. class 'torch.nn.modules.pooling.AdaptiveAvgPool2d'
\item 9. class 'torch.nn.modules.linear.Linear'
\end{itemize}
\parskip1em 
In the main text, layer numbers 0-3 are grouped and described as \textit{Resnet Conv + BN+ ReLU + MaxPool}. Layer numbers 4,5,6,7 are shown as Resnet L1,L2,L3,L4 respectively.

Expansion block consists of (1) 2D convolution with 128/64 input/output channels with kernel size \(k=3\). Stride size of 2 is used for expanding feature map size. (2) batch normalization. (3) ReLU activation.

Convolution pipe consists of (1) a 2D convolution with \(k=3\) (2) ReLU (3) a 2D transpose convolution (like deconvolution) with \(k=3\). (4) anothe ReLU. They readjust then channel sizes on the input and outputs. For example, channel adjustment layer convert 1D image channel to 3D, while cyg and cys take feature maps with 128 channels and output 4 and 8 channels respectively. Cyg's 4 channels correspond to 3 defined generators and 1 empty generator (see \(G0\) matrix). Cys' 8 channels correspond to 8 directions allowed by BSG's permutation.

\section{Training}

Training process is divided into three stages, 1. training (TS) 2. fine-tuning 1 (FT1) and fine-tuning 2 (FT2). Initial settings are shown in table \ref{tab:training_setting}. The three losses for \(y,y_g,y_s\) used are cross-entropy (CE) losses, and optimization is performed w.r.t \(total\ loss=CE_{y}+\lambda_1 CE_{y_g} +\lambda_2 CE_{y_s}\) plus parameter regularization (in pytorch, the parameters are regularized by weight decay in adam optimizer).

NVIDIA DGX-1 Deep Learning System is used for TS and FT1. 

In TS, target losses for \(y,y_g,y_s\) are considered achieved if they are achieved in that particular iteration. By contrast, in FT1, they are considered achieved only if for each of the three outputs, the average loss of the past 8 iterations achieves the set target. TS also has soft target losses for \(y,y_g,y_s\), respectively \(10^{-5},10^{-3},10^{-3}\). Once achieved, the learning rate is reduced to from \(10^{-3**}\) to \(10^{-4}\). Targets are quickly achieved, so the training took only 34.8 minutes.

In FT1, the main change from TS is lower learning rate and longer training to achieve the specified target losses. However, our GPTNet does not achieve the strict targets we set. Training took 5.2 hours.

FT2 has more manual, finely chosen parameters. The model will be saved at the iteration when the best running losses are achieved. FT2 is done in a local GPU with weaker specifications (NVIDIA GeForce GTX 1050), though it took less than 1 minute to achieve the desired output based on visual inspection of generated images of difficult classes \(c=8,9\). 

\begin{table}[h!]
\caption{Initial settings for three training phase TS, first fine-tuning F1 and second fine-tuning F2.}
\begin{center}
\begin{tabular}{cccc}
    \hline
    &TS&FT1&FT2\\    
    \hline
	n epoch&10&10&1\\
	n iterations per epoch&12800&12800&64\\
	batch size&16&16&16\\
    \hline
	\textbf{adam optimizer}&&&\\
    \hline
	learning rate&\(10^{-3**}\)&\(4\times 10^{-5}\)&\(10^{-5}\)\\
    \(\beta_1\,\beta_2\)&\((0.5,0.999)\)&\((0.5,0.999)\)&\((0.9,0.9)\)\\    
    weight decay&\(10^{-5}\)&\(10^{-5}\)&\(10^{-2}\)\\ 
    
    \hline
    \textbf{targets}&&&\\     
    \hline
    target loss \(y\)&\(10^{-5}\)&\(10^{-5}\)&\(2\times 10^{-4}\)\\
    target loss \(y_g\)&\(10^{-4}\)&\(2\times 10^{-4}\)&\(2\times 10^{-4}\)\\
    target loss \(y_s\)&\(10^{-4}\)&\(2\times 10^{-4}\)&\(2\times 10^{-4}\)\\
    target achieved at iter&13423& N.A. & N.A.\\
    \hline  
    \textbf{loss}&&&\\ 
    \hline      
    \(\lambda\) &\((1.,1.)\)&\((1.,1.)\)&\((1.,100.)\)\\ 
    \hline
    \textbf{Others}&&&\\  
    \hline
    net's iter* at phase's end&13423& 141423&141453\\    
    \hline
    &&&\\

\end{tabular}
\end{center}
\label{tab:training_setting}
\end{table}

*The GPTNet model stores the persistent total number of iterations across training phases as \textit{net's iter}.

Note that target losses are selected based on our observations of notebooks/tutorial\textunderscore gpt\textunderscore mnist\textunderscore losses.ipynb

\section{DEVELOP function}
The DEVELOP function is based on section 4.2 of the textbook whose \textit{dynamic rules} are mentioned in the main text. See the python codes directly, although the following may be helpful. Here, we will provide explanations and illustrations instead.

\textit{Indexing subtleties}: APL codes use 1-based indexing. Whereas we see from the main text that \(G0=\{\alpha:\alpha=0,1,2,3\}\), i.e. \(\alpha\) notation uses 0-based indexing, DEVELOP and related functions will point to generators in \(G0\) and \(G\) using 1-based indexing. Python arrays are 0-based too, and our repo has adjusted it accordingly. Unfortunately, bond values are directly related to the row index, hence we will see \(\pm 1\) discrepancy that just needs to be handled carefully. Otherwise, there is no technical difference.

Let us first see the results obtained by running the python version of DEVELOP defined in the textbook (which uses APL language). Run gpt/src/APL\textunderscore to\textunderscore python\textunderscore simple\textunderscore algo\textunderscore section4.2.2.py with argument 0 to 9 for \texttt{-{}-}example\textunderscore n. You will see patterns in the textbook being reproduced. For example,the following can be generated.

\begin{multicols}{5}
\begin{verbatim}
EEEEDEEEE
    C
    B
    A
    B
    C
EEEEDEEEE
\end{verbatim}
\columnbreak
\begin{verbatim}
   C
  CE
 CE
CE
A
BD
 BD
  BD
   B
\end{verbatim}
\columnbreak
\begin{verbatim}
    B
   BBB
  BBBBB
 BBBBBBB
BBBBABBBB
 BBBBBBB
  BBBBB
   BBB
    B
\end{verbatim}
\columnbreak
\begin{verbatim}
    C
    B
  DDCDD
    B
DDDDCDDDD
    B
    A
    B
DDDDCDDDD
    B
  DDCDD
    B
    C
\end{verbatim}
\columnbreak
\begin{verbatim}
      A
      A
    AAAAA
      A
  A  AAA  A
  A A A A A
AAAAAAAAAAAAA
  A A A A A
  A  AAA  A
      A
    AAAAA
      A
      A
\end{verbatim}

\end{multicols}

The original DEVELOP function consists of COMPENV and GROWTH. GROWTH1 is the specific implementation of GROWTH function. The following is the \textit{dynamic rule} for GROWTH1 that governs the change of generator configurations:
\parskip0em 
\begin{enumerate}
\item New generators can only be added to sites with empty generators \(\alpha=0\).
\item A new generator is added only if one neighboring cell competes for it i.e. if only a single non-zero bond value in that cell is contributed by the surrounding cell.
\item If a new generator is added and the single bond value is \(g+1\), then the new generator placed is \(\alpha=g\). Furthermore, the direction it is facing is the direction along which the new generator is grown from the original.
\end{enumerate}
Note that in the main text, a modified version of this is used. See the repo. Here, we illustrate using the original version as it is simpler.
\parskip1em 

\begin{figure*}[h!]
\centering
  \includegraphics[width=0.75\textwidth , trim = {0 0 0 0}]{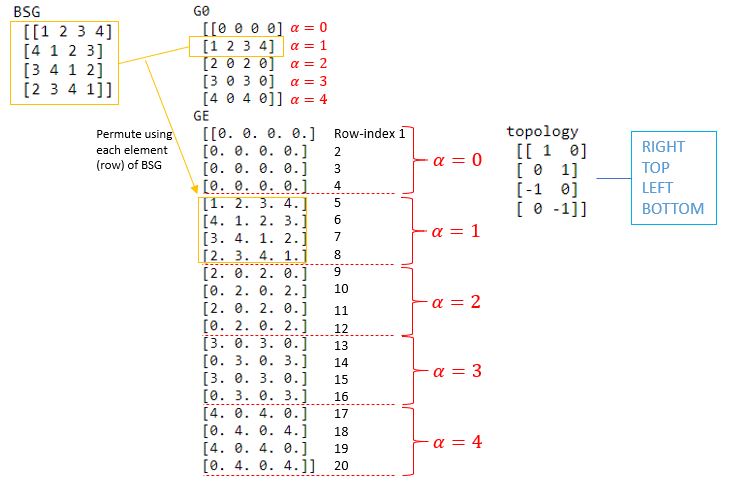}
  \caption{}
  \label{fig:setups}
\end{figure*}

To demonstrate the details of computations, we use the example in gpt/notebooks/tutorial\textunderscore getting\textunderscore started0002.ipynb ``illustrative example". The \(BSG, G0, J\) setup is shown in fig. \ref{fig:setups}. As mentioned in the main text, the figure shows how \textit{generator space} \(G\) is generated from \(G0\) and \(BSG\). The topology used consists of 4 directly adjacent neighbors in the following order: right, top, left, bottom. \(BSG\) also shows the cyclic permutation for 4 bonds attached to the generators used in this example. In summary, \(G\) consists all transformed versions of \(G0\).

\begin{figure*}[h!]
\centering
  \includegraphics[width=1\textwidth , trim = {0 0 0 0}]{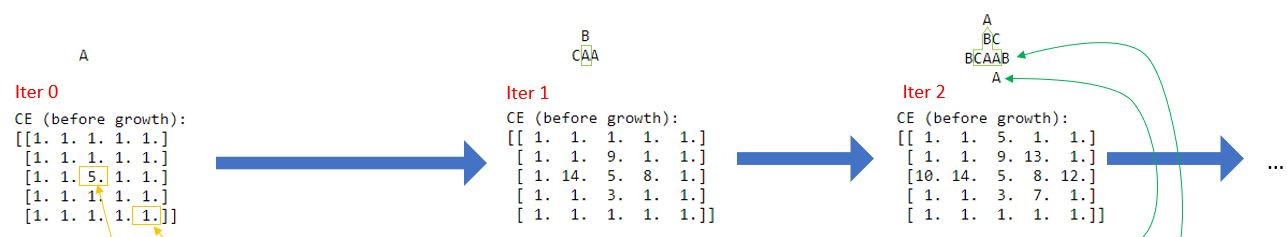}
  \caption{Observing the growth of lattice; this figure is abbreviated as we will use the more illustrative version soon.}
  \label{fig:develop}
\end{figure*}

\begin{figure*}[h!]
\centering
  \includegraphics[width=1\textwidth , trim = {0 0 0 0}]{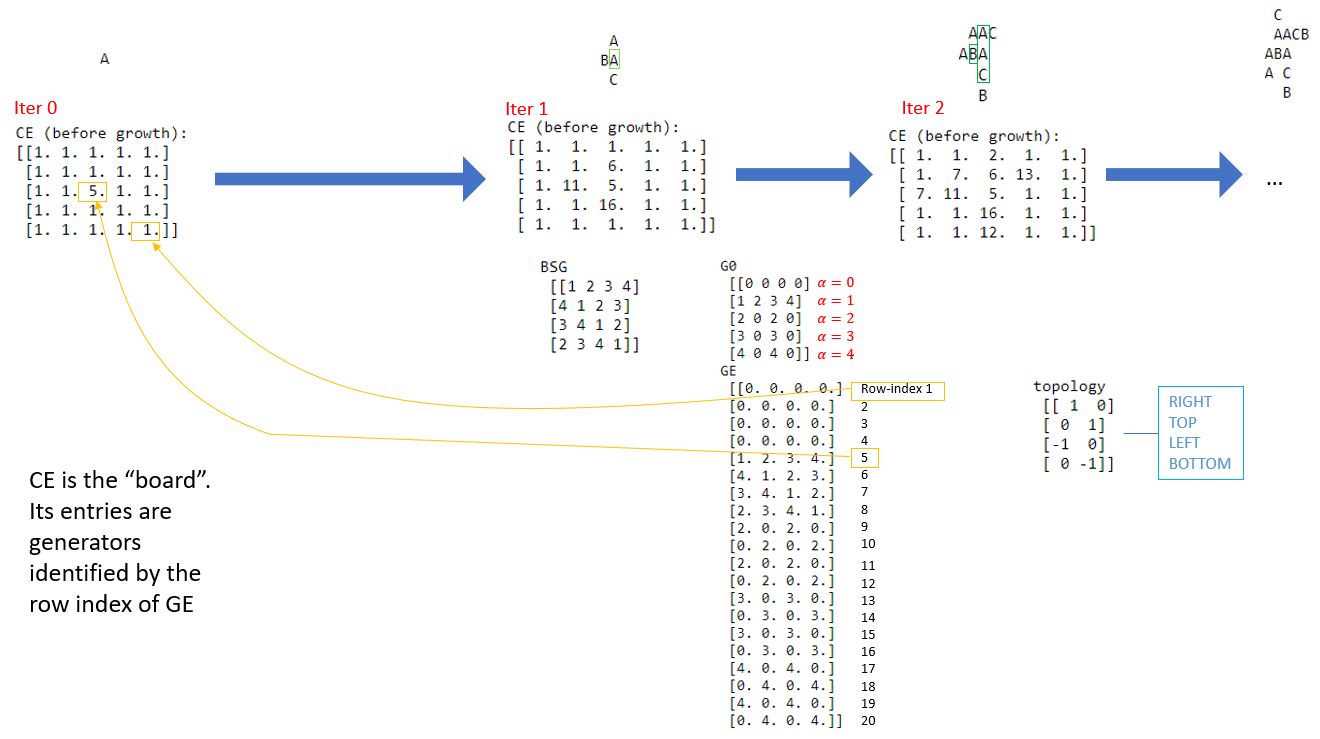}
  \caption{Revised mode of develop function is used here. The letter  representation is vailable, showing A,B,C for \(\alpha=1,2,3\) respectively. \(\alpha=0\) appears as blank.}
  \label{fig:develop_revised}
\end{figure*}

Now, create an initial configuration \(c_0\) represented by matrix \(CE\) with \(5\) at the center of \(5\times 5\) lattice. 5 is \(\alpha=1\) in its original orientation; see from fig. \ref{fig:setups} that 5 corresponds to the 5-th row of \(GE\). The rest are \(\alpha=0\) in original orientation. Fig. \ref{fig:develop} shows two steps of this configuration being developed. The function GROWTH1 works as expected. However, as in the figure, the direction of newly placed generators due to 5 are not very intuitive (we see BOTTOM/RIGHT/UP/LEFT). It actually makes sense with respect to the directions of array index (left to right, top to bottom), but we want TOP/BOTTOM/LEFT/RIGHT that corresponds to what we exactly see. Hence, we provide the `revised' mode for GROWTH1, demonstrated in gpt/notebooks/tutorial\textunderscore getting\textunderscore started0002.ipynb ``illustrative example Revised". This is shown in fig. \ref{fig:develop_revised}. Since its bond values are \(1,2,3,4\) arranged in the direction specified by the topology matrix \(J\) (RIGHT, TOP, LEFT, BOTTOM), we see that 1,6,11 and 16 are added in that order. They are respectively \(\alpha=0,1,2,3\) in the respective orientations.

\begin{figure*}[h!]
\centering
  \includegraphics[width=1\textwidth , trim = {0 0 0 0}]{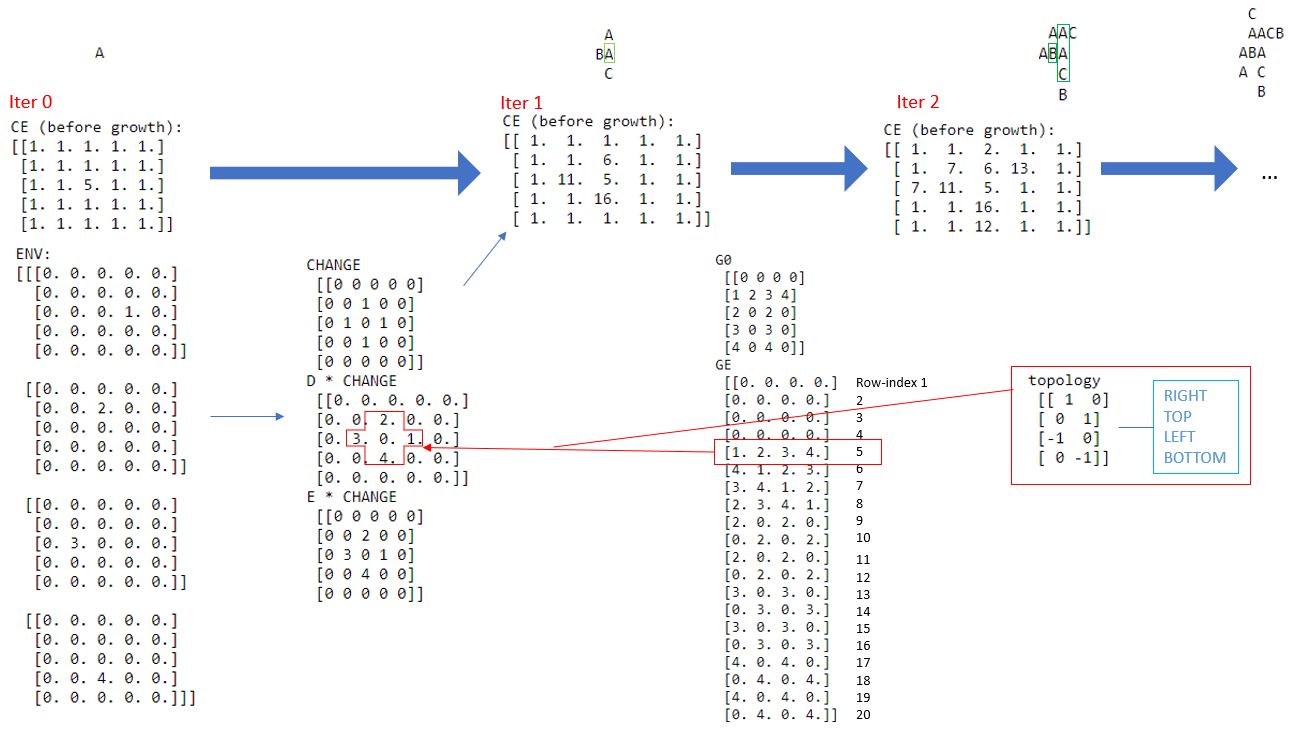}
  \caption{Inner computation of revised mode. * is entry-wise multiplication. D*CHANGE contains the final bond values, E*CHANGE contains the orientation transformations to the sites where new generators are going to be placed.}
  \label{fig:develop_inner}
\end{figure*}

Finally, to show in more details how the revised develop function is computed, see fig. \ref{fig:develop_inner}. \(ENV\) is computed by COMPENV, where each sub array corresponds to RIGHT TOP LEFT BOTTOM and the bond values 1,2,3,4 in ENV array are due to the initial element 5. CHANGE is computed according to rule 1 and 2. Indeed, the neighbors of 5 are all empty, so we can place the bond values there. At each such location, only 1 bond value is competing for the slot (all due to 5), so we can proceed by collapsing \(ENV\) into \(D\), the final bond values for this step's processing. \(E\) will contain the transformation elements (which are rotations) according to their the relative position of the new generators to the parent generators. What we see in the resulting configuration (\(CE\) in iter 1), is obtained according to rule 3. And then the process repeats. We also display the generators by letters ABCD, such that \(\alpha=0\) is blank, \(\alpha=1,2,3,4\) are A,B,C,D respectively. Notice that from iter 1 to 2, B and C will compete for the bottom left site. When this happens, rule 2 does not hold, so no new generator is placed.

\section{More Heatmaps}
\begin{figure*}[h!]
\centering
  \includegraphics[width=1\textwidth , trim = {0 0 0 0}]{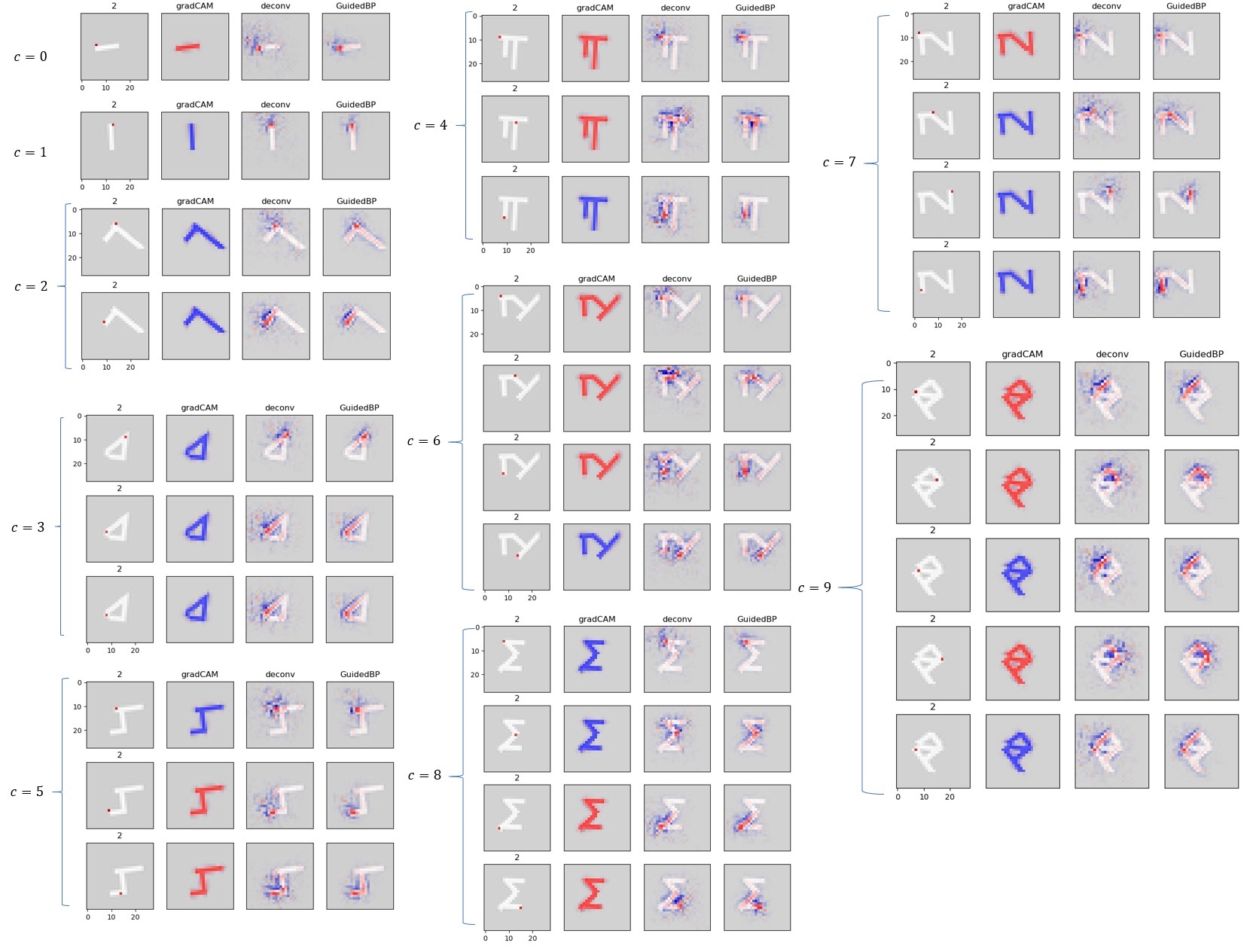}
  \caption{Sample heatmaps derived from pixel-wise \(y_g\) prediction for each class.}
  \label{fig:heatmapsyg}
\end{figure*}

\begin{figure*}[h!]
\centering
  \includegraphics[width=1\textwidth , trim = {0 0 0 0}]{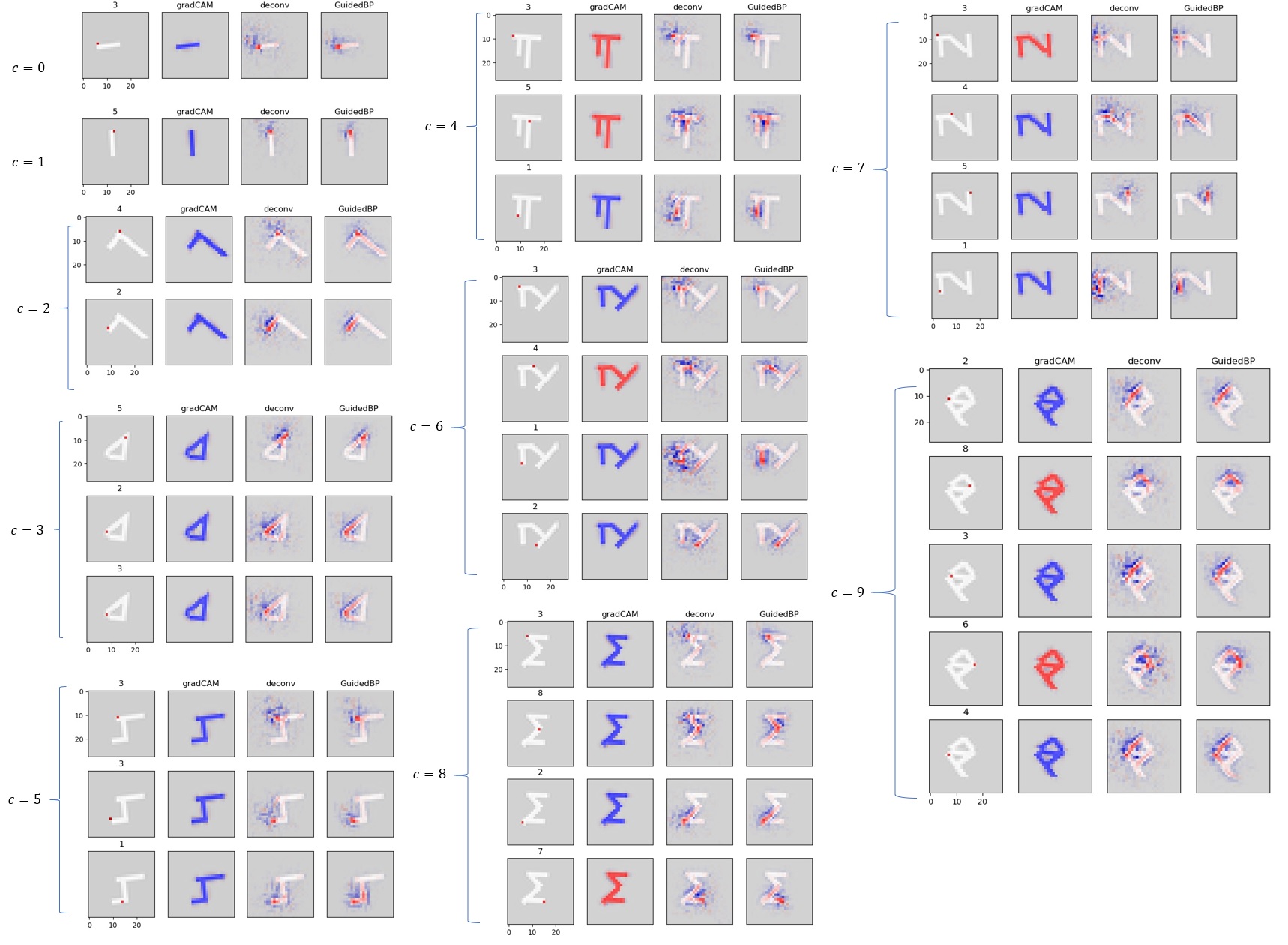}
  \caption{Sample heatmaps derived from pixel-wise \(y_s\) prediction for each class.}
  \label{fig:heatmapsys}
\end{figure*}

More heatmaps can be seen in fig. \ref{fig:heatmapsyg} for \(y_g\) derived heatmaps and fig. \ref{fig:heatmapsys} for \(y_s\) derived heatmaps.

\end{document}